%% file: main.tex
\pdfoutput=1
\documentclass{article}
\usepackage{preprint,times}

\usepackage{url}
\usepackage{graphicx}
\usepackage{amsmath}
\usepackage{amssymb}
\usepackage{booktabs}
\usepackage{multirow}
\usepackage[table,dvipsnames]{xcolor}
\usepackage{pifont}
\usepackage{microtype}
\usepackage{enumitem}
\usepackage{hyperref}

\hypersetup{
    pdftitle={Binding Phoneme Identity to Position in Neural Models and the Human Brain},
    colorlinks=true,
    linkcolor=Maroon,
    citecolor=NavyBlue,
    urlcolor=MidnightBlue
}

\newcommand{\onion}{\textsc{Onion}}
\newcommand{\decay}{\textsc{Decay}}
\newcommand{\spiral}{\textsc{Spiral}}
\newcommand{\subspace}{\textsc{Subspace}}

\title{Binding Phoneme Identity to Position in Neural Models and the Human Brain}

\author{Nafiseh Momeni\\
Université Paris Cité\\
\texttt{momeni.nafiseh@gmail.com} \\
\And
Malo Renaudin\\
École Normale Supérieure -- PSL \\
\texttt{malo.renaudin@ens.psl.eu} \\
\AND
Itzhak Fried\\
University of California, Los Angeles\\
\texttt{ifried@mednet.ucla.edu} \\
\And
Yair Lakretz\\
École Normale Supérieure -- PSL \\
\texttt{yair.lakretz@gmail.com}
}

\begin{document}

\maketitle

\input{1-intro}
\input{3-methods}
\input{4-results}
\input{5-discussion}
\input{6-statements}

\bibliography{refs}
\bibliographystyle{preprint}

\input{7-appendix}

\end{document}

%% file: 1-intro.tex
\begin{abstract}
A fundamental challenge in cognitive science and neural computation is the binding problem: how neural systems associate separately represented contents with their structural roles. Compositional generalization can be achieved by factorizing representations into \emph{roles}, the structural slots of a representation (e.g., serial position or syntactic function), and \emph{fillers}, the contents that occupy them (e.g., a phoneme or a word). Because roles and fillers can be recombined freely, a system that represents them separately can represent novel combinations. How neural systems implement role--filler binding, however, remains largely unknown. One of its simplest linguistic cases is phoneme sequencing, where phoneme identity (the filler) must be bound to position (the role) so that /gum/ and /mug/ are represented differently. While previous work identified the neural encoding of single phonemes in human auditory cortex, it remains unclear how these encodings are combined into sequences. A recent proposal from natural language processing, the `onion hypothesis' \citep{csordas-etal-2024-recurrent}, offers a concrete account: the direction of the neural population state encodes token identity, while its magnitude encodes token position. However, the onion hypothesis was formulated for random token sequences; here we extend it to natural language, where successive tokens are strongly constrained, and test it in both neural models and the human brain. We first evaluated the onion hypothesis in an encoder-decoder trained to repeat English words, combining representational geometry analyses with causal interventions. Magnitude alone turns out to be incomplete: position is encoded by magnitude together with a systematic rotation of the state direction, a scheme we formalize as the `spiral hypothesis'. We then analyzed intracranial recordings from the auditory cortex of 20 participants listening to short spoken sentences. Across superior temporal gyrus electrodes, the change in population response between consecutive phonemes of a word decreases in magnitude as position advances, while the direction of that change is positively aligned within consonants and within vowels and close to orthogonal across the two classes. This supports the magnitude coding shared by both the onion and spiral accounts. Together, these results provide a geometric account of how neural systems bind phoneme identity to position, and show how hypotheses derived from neural models can yield testable predictions about human neural computation.
\footnote{Our code is public: \url{https://github.com/nafis-momeni/spiral_representation}}
\end{abstract}

\begin{figure}[!ht]
    \centering
    \includegraphics[
    page=1,
    width=\linewidth,
    trim=0cm 17.7cm 2.8cm 0cm,
    clip
]{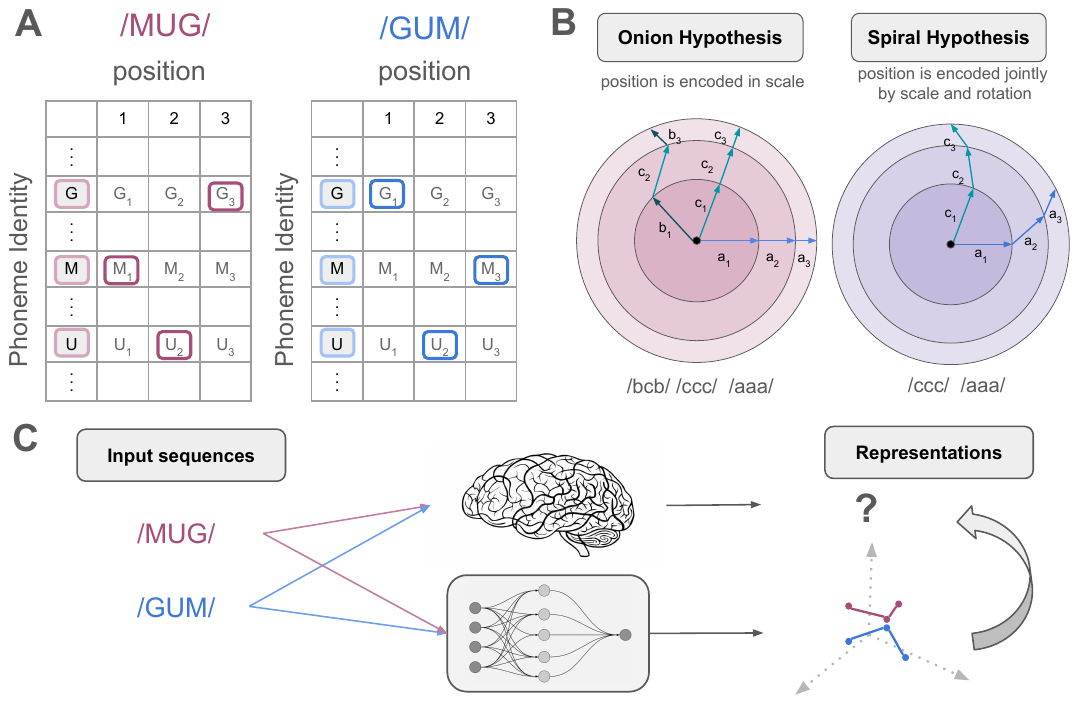}
    \caption{\textbf{Hypotheses for How a Recurrent State Binds Phoneme Identity to Position.}
    \textbf{(A)} The binding problem for sequences. Two words made of the same phonemes in a different order, /gum/ and /mug/, must map to different states, so the representation cannot depend on phoneme identity alone (e.g., a bag-of-phoneme model).
    \textbf{(B)} Two hypotheses for how position is encoded. Under the onion hypothesis (left) identity is carried by the direction of the representation and position by its magnitude, so a repeated phoneme keeps its direction across positions while its magnitude shrinks. Under the spiral hypothesis (right) identity is again carried by direction, but position is carried jointly by magnitude and rotation.
    \textbf{(C)} A recurrent network trained to repeat words serves as a hypothesis generator. Its positional representation can be both measured and causally manipulated, and the predictions it yields are then tested in human auditory-cortex recordings.}
    \label{fig:hypotheses}
\end{figure}

\section{Introduction}

Generalizing beyond the training distribution remains a central difficulty for AI models. When carefully tested, AI models still do so less systematically than humans, in, e.g., reasoning \citep{wu-etal-2024-reasoning, chollet2026arcagi2newchallengefrontier}, vision \citep{campbell2024understanding}, and language \citep{lake_baroni_2018, ismayilzada-etal-2025-evaluating}. One explanation for this gap is that humans better extract and disentangle stable structures in the data from their contents. In cognitive science, this distinction is formalized as one between \emph{roles}, the structural slots of a representation (e.g., a location in a spatial layout, or a serial position in a sequence), and \emph{fillers}, the contents that occupy them (e.g., the landmark at that location, or the phoneme at that position) \citep{smolensky1990tensor}. A system that represents roles and fillers separately can reuse a familiar role with new fillers, rather than memorizing each combination on its own \citep{ROSKIES19997bindingpb, greff2020bindingproblemartificialneural}. Such role--filler factorization is useful across many domains, such as spatial navigation \citep{whittington_beherns} and language \citep{rambaud_2026, feng_steinhardt_2023}. How neural systems implement this factorization, and the binding it requires, remains an open question.

Here we take the neural encoding of phoneme sequences as a case study. It is simple enough to make the geometry measurable: the phoneme inventory is small and discrete, and words are only a few phonemes long. Furthermore, the encoding of individual phonemes is already well characterized, in human superior temporal gyrus (STG) as combinations of phonetic features \citep{mesgarani_phonetic_2014, Yi_2019}. This prior work characterizes the encoding of identity alone, however, leaving open how it is bound to a position.

To address this question, we build on a hypothesis recently proposed in natural language processing for recurrent neural networks (RNNs). RNNs were inspired by cortical recurrence \citep{ELMAN1990, hopfield_1982}. Unlike transformers, which attend over a stored history of past inputs, RNNs update a fixed-size state as each input arrives, a mechanism closer to how the brain processes sequences. This also gives them a computational advantage: they process a sequence in linear time and at constant memory per step, avoiding the growing cache that attention requires. For this reason recurrent architectures have come back into wide use in the form of state-space models \citep{gu_2022, Gu2023MambaLS}. In such recursive architecture, identity and position must be jointly encoded in a single evolving state, unlike in transformers, where position is supplied by a separate positional embedding. The \textit{Onion Hypothesis} gives a concrete account of how this is done: the direction of the state encodes token identity and its magnitude encodes token position, with earlier tokens held at larger magnitude \citep{csordas-etal-2024-recurrent}. However, the onion hypothesis was formulated for random token sequences, whereas natural language is strongly constrained by phonotactics \citep{VITEVITCH1999_phonotactics, Leonard_chang_2015_prob}, consonant--vowel alternation \citep{Clements_1990_CV}, and frequency \citep{piantadosi_2014_zipf}. This leaves two open questions: does the onion hypothesis hold for natural language, and does it capture how the human brain binds phoneme identity to position?

We first evaluated the Onion Hypothesis in a pre-trained encoder-decoder RNN trained to repeat English words at natural frequencies \citep{dager2025neuralmodelwordrepetition}. A hypothesis-free analysis of the encoder geometry reproduced the predicted magnitude decay, but showed that the direction of a phoneme's contribution is not accurately preserved across positions, which the hypothesis requires. We therefore propose a refinement, the \textit{Spiral Hypothesis}, in which position is carried jointly by a decaying magnitude and a systematic rotation of the state direction. To compare the two accounts we turned to causal interventions on the encoder state, editing it as each hypothesis prescribes and asking whether the decoder then produces the word that hypothesis predicts. The spiral model outperformed the onion model and also a subspace baseline. Furthermore, a control experiment confirmed that its advantage comes from the rotation itself rather than from its additional parameters. Interventions also failed more often when the edit would produce a phonotactically illegal word, showing that the positional code is shaped by the statistics of the language -- an effect that random sequences cannot reveal.

We then asked whether this geometry is present in the human brain, analyzing intracranial recordings from 20 participants listening to short spoken sentences. Across superior temporal gyrus electrodes, the change in population response between consecutive phonemes decreases in magnitude as position advances, as both hypotheses predict. This decline persists when phoneme predictability is controlled, indicating that magnitude encodes position rather than surprisal. The direction of the change is aligned within consonants and within vowels and close to orthogonal across the two classes, consistent with direction encoding phoneme identity. Together, these results give a geometric account of how phoneme identity is bound to position, and show how a hypothesis developed for neural language models can yield testable predictions about the human brain.

%% file: 3-methods.tex
\section{Experimental Setup}
\label{sec:methods}

\subsection{Recurrent Model and Contextualized Phoneme Embeddings}
\label{sec:model}

\paragraph{Pretrained Model.}
We study the pretrained Neural Word Repetition (NWR) model of \citet{dager2025neuralmodelwordrepetition}. It is an encoder--decoder network with LSTM units \citep{Hochreiter_LSTM}, trained on a word repetition task. The encoder and decoder each have a single hidden layer of size $d = 128$, and the input and output vocabulary consists of 42 phonemes. The model was trained on the $30{,}000$ most frequent English words, sampled according to their frequency. Its hyperparameters ($d$, number of layers, etc.) were chosen as the smallest model that achieved perfect repetition on the training words. After training, the NWR generalizes to unseen pseudowords (phoneme sequences outside the training vocabulary that follow English phonotactics), repeating them with $97.25\%$ accuracy \citep{dager2025neuralmodelwordrepetition}.

\begin{figure}[!t]
    \centering
    \includegraphics[
    page=3,
    width=\linewidth,
    trim=0cm 23.5cm 0cm 0cm,
    clip
]{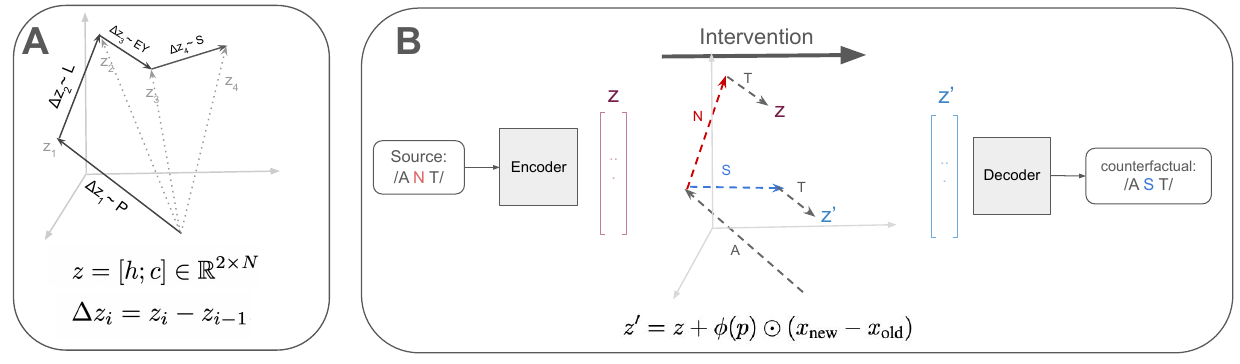}
   \caption{\textbf{Delta states and the intervention.}
   \textbf{(A)} How the contribution $\Delta z_i$ of the phoneme at position $i$ is computed from consecutive encoder states.
   \textbf{(B)} Each phoneme adds an increment $\Delta z_i$, and the intervention maps the increment of the old phoneme /N/ to that of the new phoneme /S/, turning \texttt{ANT} into \texttt{AST}.}
    \label{fig:delta}
\end{figure}

\paragraph{Contextualized Phoneme Embeddings.}
We take the encoder state to be the concatenation $z = [\,h; c\,] \in \mathbb{R}^{2d}$ of the hidden and the cell state, since the information the decoder reads may be distributed across both (Appendix~\ref{app:state} reports the analyses on each component separately). To achieve strong repetition performance, $z$ has to encode every preceding phoneme identity together with its position. To isolate the representation of a single phoneme in the sequence, we compute
\begin{equation}
    \Delta z_i \;=\; z_i - z_{i-1}, \qquad z_i = \mathrm{Encoder}(w_{1:i}),
    \label{eq:deltaz}
\end{equation}
for every word $w$ (Figure~\ref{fig:delta}A). Since $z_i$ accumulates the encoding of all phonemes up to position $i$, $\Delta z_i$ isolates the update induced by the phoneme at position $i$, given the preceding context. We therefore refer to $\Delta z_i$ as the \emph{contextualized embedding} of that phoneme, in contrast to the non-contextualized input embedding, which depends on phoneme identity alone. The onion hypothesis makes two predictions about these embeddings: their norm should decrease with position, and their direction should carry phoneme identity and therefore be preserved when the same phoneme occurs at a different position.

\paragraph{Representational Analyses.}
Two descriptive measures test these predictions. For the positional prediction, we take the $\ell_2$ norm of $\Delta z_i$ as a function of position. For the identity prediction, we compute the cosine similarity between pairs of contextualized embeddings at two levels of granularity. At the fine level, pairs are grouped by whether they share the same phoneme identity; if direction encodes identity, embeddings of the same phoneme should be more aligned than those of different phonemes. At the coarse level, pairs are grouped by broad phonetic class (consonant or vowel); if direction encodes identity, this structure should carry over to classes, with higher alignment within a class than across classes. The same measures are applied to the human recordings in Section~\ref{sec:ieeg}, where the identity prediction is tested at the coarse level only.

\subsection{Causal Interventions}
\label{sec:intervention}

Representational analyses and descriptive geometry show how representations are organized but cannot establish whether a geometric property is used when the model produces an output. For causal claims, we adopt the causal abstraction framework \citep{Geiger2021}: we edit the encoder state so that it should correspond to a sequence in which one phoneme has been replaced, and check whether the frozen decoder generates that counterfactual sequence.

\paragraph{Counterfactual Data.}
\label{sec:data}
Counterfactual pairs are built from the $30{,}000$ training words by choosing a position at random and replacing its phoneme with one sampled from the vocabulary. Each example thus pairs a real word with a counterfactual that differs from it at a single position $p$ (Appendix Table~\ref{tab:examples}). The encoder reads one member of the pair, the \emph{base}, and the decoder must produce the other, the \emph{target}. We run both directions: taking the counterfactual as target requires the intervention to write an unattested phoneme sequence into the state, and taking the real word as target requires it to undo that edit. We verified that all base and targets are repeated correctly by the frozen model before any intervention, so a failure reflects the edit rather than the model. Because random substitution often yields sequences that no English word could potentially contain, we also define two phonotactic filters over the counterfactuals: one enforces the consonant--vowel patterns seen in training, and the other rejects phoneme bigrams or trigrams absent from the training words. Appendix~\ref{app:phonotactics} describes the filters and varies them as an experimental factor.

\paragraph{Additive Intervention.}
Following \citet{csordas-etal-2024-recurrent}, we write the change of identity into the state by adding a vector to it. Let $x_{\text{old}}$ and $x_{\text{new}}$ be the embeddings of the base and target phonemes at position $p$. A learned \emph{positional scaling} vector $\phi(p) \in \mathbb{R}^{2d}$ controls how the change is written into the state,
\begin{equation}
z' = z + \phi(p) \odot (x_{\text{new}} - x_{\text{old}}),
\label{eq:intervention}
\end{equation}
where $\odot$ is elementwise multiplication. The embedding difference sets the direction of the change, and $\phi(p)$ sets how strongly and along which axes that change enters the state at position $p$ (Figure~\ref{fig:delta}B). The edited state $z'$ is passed to the frozen decoder, and $\phi$ is trained by minimizing the cross-entropy between the target sequence and the decoder output, generated without teacher forcing. We report exact-sequence match throughout.

\paragraph{Identity Embeddings.}
The embeddings $x_{\text{old}}$ and $x_{\text{new}}$ are initialized in one of three ways: (i) from the input non-contextualized embedding layer of the pretrained model, (ii) from the mean contextualized embedding $\Delta z$ of each phoneme over the training set (Section~\ref{sec:model}), or (iii) randomly (i.e., learned from scratch). For (i) and (ii), we evaluate fixed embeddings (i.e., no learning) and fine-tuned embeddings, jointly with $\phi$.

\paragraph{Positional scaling functions.} We parameterize the positional scaling $\phi(p)$ as a function of position, $\phi(p) = f_\theta(p)$, with parameters $\theta$ shared across positions. Each choice of $f_\theta$ is a structural hypothesis about how position enters the state; because $f_\theta$ is defined for every $p$, it also predicts the scaling at positions on which it was not fitted.  We compare three scaling functions, with $8 \times d$, $6 \times d$, and $7 \times d$ learnable parameters, respectively. The first reproduces the original onion account, the second isolates its core claim of magnitude decay, and the third adds a rotation of the identity direction with position. Powers of vectors are taken elementwise.

\begin{enumerate}
  \renewcommand{\labelenumi}{(\roman{enumi})}
  \item \textbf{\onion{}.} As a reference, we use the scaling function of \citet{csordas-etal-2024-recurrent}, an exponential decay with a linear drift:
  \begin{equation}
    \phi(p) = g \odot \gamma^{p} + \beta\,p + b, \qquad g, \gamma, \beta, b \in \mathbb{R}^{2d}.
  \end{equation}

  \item \textbf{\decay{}.} The linear term in \onion{} is unbounded and can grow with position, which departs from the hypothesis that magnitude decreases with position. \decay{} removes this term and constrains the scaling to decay monotonically:
  \begin{equation}
    \phi(p) = s(p) = \operatorname{softplus}(g) \odot \sigma(\gamma)^{p} + b, \qquad g, \gamma, b \in \mathbb{R}^{2d},
    \label{eq:decay}
  \end{equation}
  where $\sigma$ is the logistic sigmoid. Since $\sigma(\gamma) \in (0,1)^{2d}$ and $\operatorname{softplus}(g) > 0$, each dimension decays monotonically to its baseline $b$. \decay{} thus tests whether magnitude decay alone is sufficient to write identity at the correct position.

  \item \textbf{\spiral{}.} The descriptive analyses (Section~\ref{sec:results-descriptive}) show that contextualized embeddings of the same phoneme become less aligned as their positions grow further apart, suggesting that the identity direction rotates with position. In addition, because \decay{} converges to a baseline, it separates late positions only weakly, and a rotation provides a second positional signal that does not saturate. \spiral{} therefore applies a position-dependent rotation to the \decay{} scaling, $\phi(p) = R(p)\,s(p)$, with $s(p)$ as in Eq.~\eqref{eq:decay}. We build $R(p)$ with the geometric construction of rotary position embeddings \citep{su_roformer:_2024}, borrowing the construction but not its use as a transformer position encoding. The $2d$ dimensions are grouped into $d$ pairs, and the plane spanned by each pair is rotated by an angle that grows linearly with position:
  \begin{equation}
    R(p) = \operatorname{diag}\big(R_1(p), \dots, R_d(p)\big), \quad
    R_i(p) = \begin{pmatrix} \cos\theta_i & -\sin\theta_i \\ \sin\theta_i & \cos\theta_i \end{pmatrix}, \quad
    \theta_i = p\,\omega_i.
  \end{equation}
  The only added parameters are the $d$ frequencies $\omega_i$, so different parts of the state can rotate at different rates. Identity is still written along a direction, but that direction turns linearly with position while its scale decays exponentially.
\end{enumerate}

Unconstrained variants that fit the scaling at each position independently, and thus bound what the additive edit can achieve, are reported in Appendix~\ref{app:methods}.

\paragraph{Subspace baseline.}
The three scaling functions assume that position is encoded in the magnitude and identity in the direction of a shared state. A common alternative is that each position occupies its own linear subspace of the state. \subspace{} tests this alternative using a variant of distributed alignment search (DAS; \citealp{geiger_das_2024}), following \citet{csordas-etal-2024-recurrent}. Instead of adding a vector, \subspace{} overwrites part of the base state with content copied from a third sequence, the \emph{source}, which already carries the target phoneme. It learns an orthogonal rotation of $z$ into a basis in which the positional subspaces are axis-aligned, together with an assignment of each rotated dimension to a positional variable spanning $k$ consecutive positions. The source agrees with the target only on the positions covered by the intervened variables and is resampled uniformly elsewhere, so it provides no context that the additive methods lack. We report $k \in \{1,2\}$; the construction is given in Appendix~\ref{app:das}.

\subsection{Human Intracranial Recordings}
\label{sec:ieeg}

\paragraph{Data and preprocessing.}
We analyze high-gamma activity (70--150~Hz) recorded from 59 sound-responsive superior temporal gyrus (STG) electrodes, identified in 9 out of the 20 participants of the intracranial dataset of \citet{Lakretz_2026}, while they listened to 152 short spoken sentences (508 words, 1618 phonemes). Details are given in Appendix~\ref{app:ieeg}.

\paragraph{Contextualized neural responses.}
Phoneme boundaries were obtained by forced alignment \citep{MFA} with manual correction. Let $\bar{r}(p) \in \mathbb{R}^{59}$ be the high-gamma amplitude of the selected electrodes, averaged over the interval of phoneme $p$. By analogy with Eq.~\eqref{eq:deltaz}, we define the contextualized neural response to the $i$-th phoneme of a word as
\begin{equation}
    \Delta_i \;=\; \bar{r}(p_i) \;-\; \bar{r}(p_{i-1}).
    \label{eq:delta}
\end{equation}
This captures the contribution of each successive phoneme rather than the cumulative response.
We excluded word-initial phonemes, for which $\Delta_i$ is undefined, and the first word of each sentence, which carries the sentence-onset response. This leaves 1110 responses. The positional prediction is tested by relating $\|\Delta_i\|$ to within-word position.

\paragraph{Controlling for predictability.}
Later phonemes in a word are on average more predictable, and less predictable phonemes elicit larger cortical responses \citep{brodbeck_2018, donhauser_2020}. A norm that declines with position could therefore reflect declining surprisal. To separate the two, we compute the cohort surprisal of each phoneme, $s(p_i) = -\log_2 P(p_i \mid p_1 \ldots p_{i-1})$, over the entries of the CMU Pronouncing Dictionary \citep{cmudict} (details in Appendix~\ref{app:ieeg}).

\paragraph{Identity analysis.}
For the identity prediction, we compute the cosine similarity between all pairs of responses and group the pairs into consonant--consonant, vowel--vowel, and consonant--vowel. We quantify the separation between within- and between-class similarities with the area under the ROC curve (AUC): the probability that a random within-class pair is more similar than a random between-class pair ($0.5$ is chance). As a rank statistic, AUC is insensitive to the scale and skew of the similarities and comparable across contrasts with different pair counts. Significance is assessed by permuting phoneme class labels (1000 permutations).

%% file: 4-results.tex
\section{Results}
\label{sec:results}

\begin{figure}[!t]
    \centering
    \includegraphics[
    page=2,
    width=\linewidth,
    trim=0cm 24.3cm 0.5cm 0cm,
    clip]{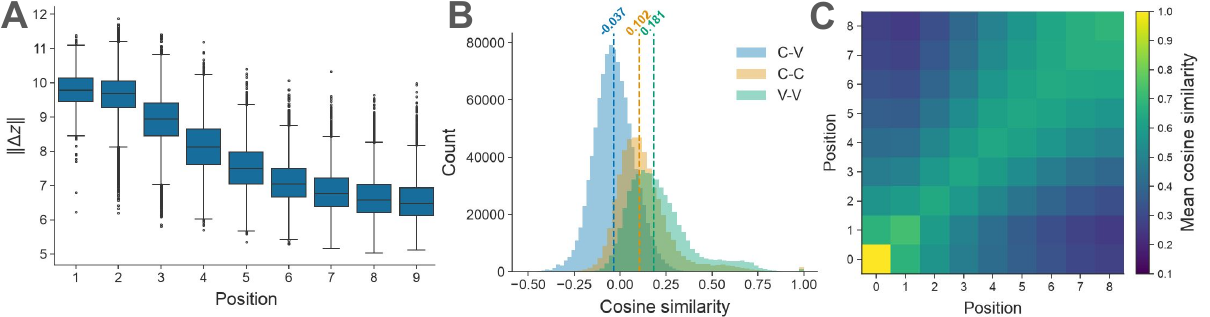}
    \caption[Hypothesis-free analysis of the geometry of phoneme-sequence encoding]{\textbf{Hypothesis-Free Analysis of the Geometry of Phoneme-Sequence Encoding.}
    \textit{Phoneme-position encoding:} \textbf{(A)} Norm of the contextualized phoneme embedding $\Delta z$ as a function of phoneme position. In accordance with the onion hypothesis, the norm is largest at the first position and decreases for later positions in the word.
    \textit{Phoneme-identity encoding:} The onion hypothesis holds that identity is carried by the direction of $\Delta z$. \textbf{(B)} Distribution of cosine similarities between $\Delta z$ vectors, split by within-class (C--C, V--V) and across-class (C--V) pairs, where C and V denote consonants and vowels; medians dashed. Within-class pairs are positively aligned, while across-class pairs are close to orthogonal, so direction carries identity information. \textbf{(C)} Mean cosine similarity between contextualized embeddings of the same phoneme at each pair of positions, averaged over the phoneme inventory. If direction were preserved across positions, as the onion hypothesis requires, all entries would be close to one. Instead, similarity varies systematically with position, indicating that the identity direction changes as position advances and motivating the spiral refinement.}
    \label{fig:descriptive}
\end{figure}

\subsection{The Onion Hypothesis Holds for Scale but Fails for Orientation}
\label{sec:results-descriptive}

We first conducted a hypothesis-free descriptive analysis, examining the magnitude and direction of phoneme encodings in order to characterize the geometry of phoneme-sequence encoding empirically. The representations reproduced the broad onion pattern. The norm of the delta state was largest at the first position and decreased with position (Figure~\ref{fig:descriptive}A), matching the magnitude decay the onion hypothesis predicts.

To test whether the organization of identity respects broader phonological structure, we grouped phonemes into the two major functional classes, consonants (C) and vowels (V). Direction did carry identity in the way the hypothesis proposes -- within-type delta states (C-C or V-V pairs) were more similar to each other than cross-type ones (C-V pairs, Figure~\ref{fig:descriptive}B).

Examining how identity is preserved across positions, however, revealed a departure from the onion account. If position were carried by magnitude alone, the contextualized embeddings of a given phoneme would point in the same direction at every position, and since cosine similarity ignores magnitude, their similarity would be close to one for any pair of positions. We therefore computed, for each phoneme, the mean cosine similarity between its embeddings at every pair of positions, and averaged over the phoneme inventory (Figure~\ref{fig:descriptive}C). Similarity is high between nearby positions but declines with the distance between them, approaching orthogonality for distant positions. The identity direction is therefore not fixed but changes systematically with position. This descriptive analysis alone cannot tell whether this change reflects a systematic rotation, as the spiral hypothesis proposes, or, for example, the contextual influence of the surrounding phonemes. The causal comparison in Section~\ref{sec:results-intervention} is designed to separate the two.

\begin{figure}[!t]
    \centering
    \includegraphics[
    page=4,
    width=\linewidth,
    trim=0cm 24cm 0cm 0cm,
    clip
]{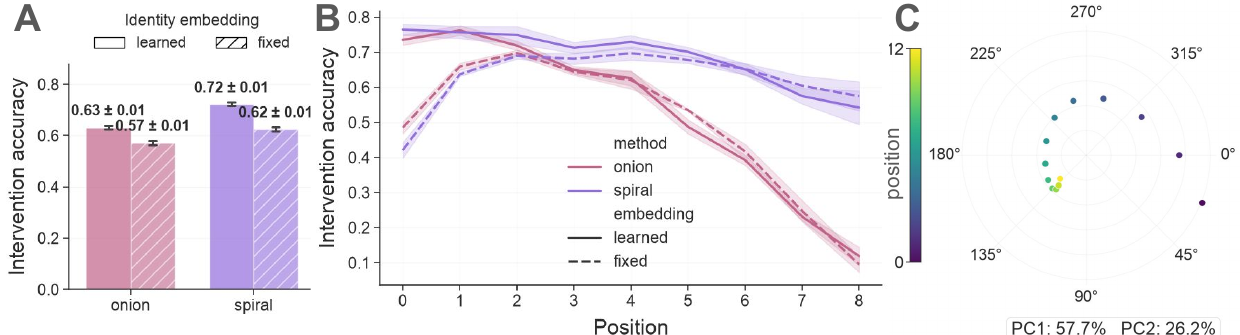}
\caption{\textbf{The Spiral Modulation Outperforms the Onion Modulation.}
\textbf{(A)} Mean intervention accuracy by method, with identity embeddings
either learned or fixed. The spiral method outperforms the onion method in both
settings.
\textbf{(B)} Intervention accuracy by position. The two methods are comparable over the first few positions, but the onion method degrades sharply from position 4 onward while the spiral method declines only gradually, indicating that the rotation matters most where magnitude decay alone has compressed the
representation.
\textbf{(C)} PCA of the learned modulation $\phi$ for the
spiral method, in polar coordinates, one point per position (colour). The
radius decays roughly exponentially with position, while the angle turns monotonically.
}
    \label{fig:intervention-results}
\end{figure}

\begin{table}[h]
\centering
\small
\setlength{\tabcolsep}{6pt}
\renewcommand{\arraystretch}{1.15}

\begin{tabular}{l c c c c}
\toprule
& \multicolumn{2}{c}{Subspace baseline} & \multicolumn{2}{c}{Scale based} \\
\cmidrule(lr){2-3} \cmidrule(lr){4-5}
Target & DAS, $k = 1$ & DAS, $k = 2$ & Onion & Spiral \\
\midrule
Pseudoword & 0.000 $\pm$ 0.000 & 0.173 $\pm$ 0.016 & 0.631 $\pm$ 0.008 & \textbf{0.723} $\pm$ 0.006 \\
Real word  & 0.102 $\pm$ 0.005 & 0.227 $\pm$ 0.005 & 0.901 $\pm$ 0.004 & \textbf{0.927} $\pm$ 0.003 \\
\bottomrule
\end{tabular}
\caption{\textbf{Causal interventions show that the spiral model outperforms the onion and subspace models.} Intervention accuracy by method and target, mean $\pm$ standard deviation over four cross-validation seeds (Section~\ref{sec:data}).}
\label{tab:results_main}
\end{table}

\begin{figure}[t]
    \centering
    \includegraphics[
    page=5,
    width=\linewidth,
    trim=0cm 17.2cm 0cm 0cm,
    clip
]{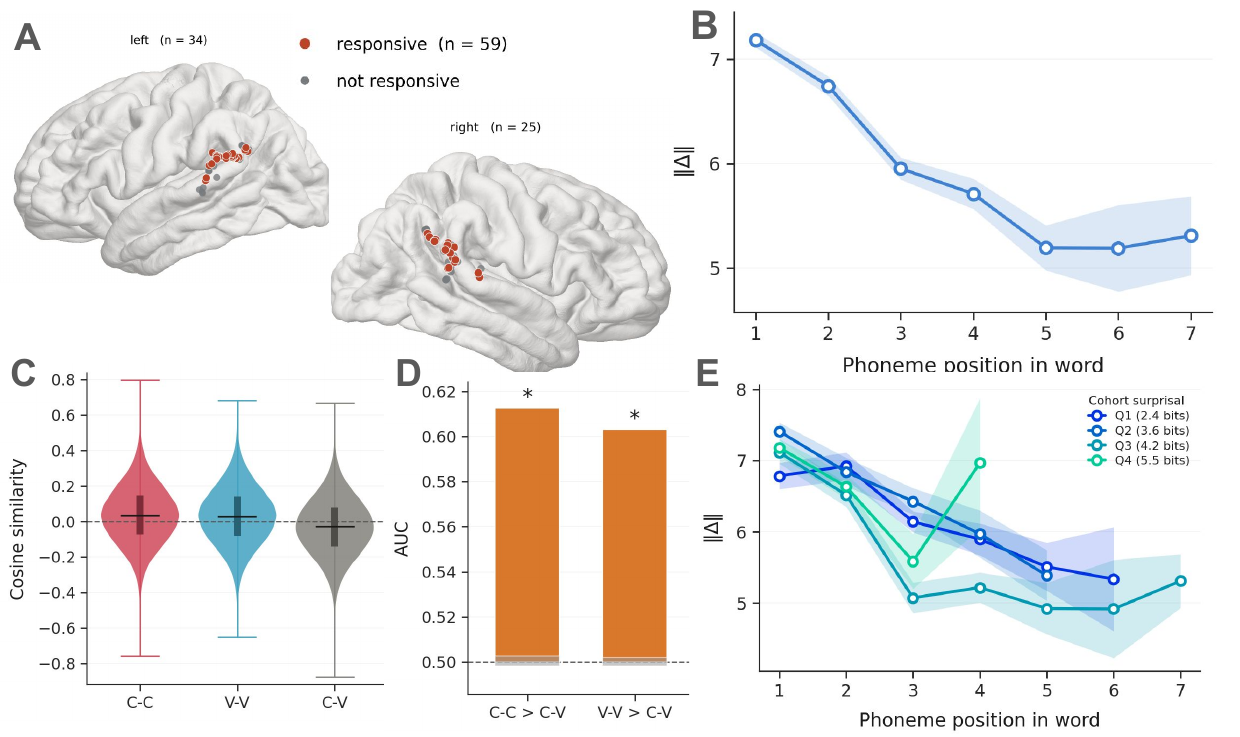}
\caption{\textbf{Phoneme representation in the human superior temporal gyrus.}
\textbf{(A)} STG electrodes on the MNI template. Red marks the 59 of 108 that respond to the speech stimuli and are used in all analyses.
\textbf{(B)} Norm of phoneme representations against its within-word position, shading is SEM.
\textbf{(C)} Cosine similarity between pairs of phoneme representations, split by consonant and vowel pair type.
\textbf{(D)} AUC for the same contrasts, the probability that a within-class pair is more similar than a between-class pair ($0.5$ is chance, dashed). Grey band is the 95\% interval of the label-permutation null (1000 shuffles), $*$ marks $p < 0.001$.
\textbf{(E)} Same as (B) but recomputed separately within each of four surprisal quartiles, showing that the positional gradient is not carried by phoneme predictability.}
    \label{fig:stg-results}
\end{figure}

\subsection{Causal State Interventions Support Spiral Encoding}
\label{sec:results-intervention}

The descriptive analysis showed that position alters the direction of the contextualized embeddings as well as their magnitude. To test whether the model uses this rotation, we compared the positional scaling functions of Section~\ref{sec:intervention} by editing the encoder state and scoring the frozen decoder on exact sequence match. We report separately the edits that turn a real word into a pseudoword and the edits in the opposite direction.

\textbf{The spiral outperforms scale-only and subspace accounts.}
\spiral{} achieved the highest accuracy in both directions, above \onion{} and far above \subspace{} (Table~\ref{tab:results_main}). Figure~\ref{fig:intervention-results}A breaks down the comparison between \spiral{} and \onion{} for edits whose target is a pseudoword, with identity embeddings either fixed or fine-tuned (Section~\ref{sec:intervention}). Fine-tuning improves accuracy for both methods (\onion{}: $0.57$ to $0.63$; \spiral{}: $0.62$ to $0.72$), but \spiral{} outperforms \onion{} in both settings, so its advantage does not depend on the added flexibility of learned embeddings. Figure~\ref{fig:intervention-results}B shows that the advantage is concentrated at late positions, where \onion{} degrades sharply while \spiral{} declines only gradually. A decaying scale alone compresses late positions together, so an additional positional signal matters most there.

\textbf{The advantage comes from the rotation, not from capacity.}
Adding the rotation to \decay{} raises accuracy from $0.64$ to $0.72$, whereas adding the same rotation to \onion{} lowers it from $0.65$ to $0.62$ (Appendix~\ref{app:rotation}). Since \spiral{} also has fewer parameters than \onion{}, its advantage cannot be attributed to additional capacity. The rotation helps only when combined with a monotonically decaying scale, which is the combination the spiral hypothesis specifies. Moreover, the additive edit receives only the position and the identities of the base and target phonemes, not the surrounding phonemes, so the benefit of rotation cannot reflect contextual influence. Projecting the learned positional scaling of \spiral{} onto its leading principal components and reading it in polar coordinates shows the geometry directly: the radius decays with position while the angle turns steadily (Figure~\ref{fig:intervention-results}C).

\textbf{Positions do not occupy separate subspaces.}
\subspace{} fails almost completely at $k = 1$. This is expected if all positions share the same directions and are distinguished by magnitude and rotation, since such a code provides no position-specific subspace to swap. \citet{csordas-etal-2024-recurrent} report the same failure for their smallest recurrent models.

\textbf{Intervention success depends on phonotactics.}
Finally, intervention success depends on the phonotactic legality of the word the edit creates. Accuracy rises from $0.722$ with no filter to $0.866$ once counterfactuals containing unattested phoneme trigrams are excluded (Appendix~\ref{app:phonotactics}). The positional code is therefore shaped by the statistics of the training language, an effect that random token sequences cannot reveal.

Together, these results show that the model uses the rotation suggested by the descriptive analysis to encode position. The positional code combines a decaying magnitude with a systematic rotation of the identity direction. It is not captured by magnitude alone or by separate positional subspaces, and it is shaped by the phonotactics of the training language.

\subsection{Human Auditory Cortex Encodes Phoneme Position in Response Magnitude}
\label{sec:results-stg}

The model analyses yield concrete, quantitative predictions about how a neural population should bind phoneme identity to position. Testing them in the human auditory cortex requires recordings that resolve individual phonemes within a word, both in time and in space, which non-invasive methods cannot provide. Intracranial recordings from patients undergoing epilepsy monitoring offer a rare opportunity to meet this requirement. We therefore asked whether the geometry found in the model is also present in the superior temporal gyrus (STG). We identified speech-responsive STG electrodes (Figure~\ref{fig:stg-results}A) in our dataset (Section~\ref{sec:ieeg}) and applied the same construction as in the model, defining neural phoneme embeddings as the change in the high-gamma population response from one phoneme to the next (Eq.~\eqref{eq:delta}). The onion and spiral accounts share two predictions: the magnitude of these embeddings should decrease with position, and their direction should carry phoneme identity.

\textbf{Magnitude decreases with position.}
As in the model, the norm of the neural phoneme embeddings decreases monotonically with the phoneme's position in the word (Figure~\ref{fig:stg-results}B; Spearman $\rho = -0.38$, $p = 1.6 \times 10^{-38}$, $n = 1110$). This is the signature predicted by both accounts. In natural speech, position might be confounded with predictability, because later phonemes in a word are on average more predictable, and predictable phonemes elicit weaker cortical responses. We therefore repeated the analysis within each of four cohort-surprisal quartiles. The decrease with position is present in every quartile (Figure~\ref{fig:stg-results}E). The only departure is an increase at position 4 in the highest-surprisal quartile. The gradient is thus not a by-product of surprisal, indicating that human auditory cortex carries a magnitude code for within-word position.

\textbf{Direction carries phoneme class.}
The second prediction is that the direction of the neural phoneme embeddings carries phoneme identity. Embeddings of phonemes from the same class are weakly but positively aligned, whereas pairs from different classes are close to orthogonal (Figure~\ref{fig:stg-results}C). The AUC analysis further confirms that within-class pairs are more similar than across-class pairs (Figure~\ref{fig:stg-results}D). Consonant--consonant (and also vowel--vowel) pairs are not reliably distinguished from each other (AUC $0.51$), as expected if the two classes occupy comparably broad regions of the space.

\textbf{An open prediction: rotation in cortex.}
In human STG, we thus find position-dependent magnitude decrease as in the onion and spiral accounts. The component that distinguishes the spiral account, a position-dependent rotation of each phoneme's direction, makes a further and more specific prediction. Testing this, however, requires many repetitions of the same phonemes at several within-word positions, which current existing data do not provide. Our study therefore provides an open prediction for future intracranial experiments.

%% file: 5-discussion.tex
\section{Discussion}
\label{sec:discussion}

We asked whether the onion hypothesis, established on random token sequences, holds for a recurrent network trained on natural words, and whether its geometry has a counterpart in the human brain. In the model, position is carried by a decaying magnitude together with a systematic rotation, which \spiral{} captures more accurately than scale-only accounts while using fewer parameters. In human STG, the magnitude component is present and is not explained by phoneme predictability.

Our findings add to a growing body of work showing that neural language models encode linguistic structure in the geometry of their representations \citep{hewitt2019structural}. A closely related polar geometry has been found for syntax \citep{diego-simon_2024, diego2025probing}, as well as for semantic structures \citep{diego2026polar}. Our results can also be interpreted in terms of role--filler binding. In the additive intervention (Eq.~\eqref{eq:intervention}), the identity embedding is the filler, the positional scaling $\phi(p)$ is the role vector, and elementwise multiplication binds them. This retains only the diagonal of their tensor product \citep{smolensky1990tensor}, so the onion and spiral accounts are both compressed tensor product representations (TPRs) that differ only in their role vectors. Phoneme--position binding is thus another case in which neural networks spontaneously learn TPRs \citep{mccoy2018rnns, soulos2020discovering, mccoy2026emergent}. Together with the polar code for syntax and semantics, this suggests a general scheme in which distinct geometric components encode distinct variables of structured representations.

The geometry also links to serial-order memory. The onion account is a primacy gradient \citep{hurlstone_2014}, and the rotation adds explicit position marking, as in oscillator models \citep{brown_preece_hulme_2000}. It also explains the serial-position curve: the first phoneme is written at the largest scale, making it robust to interference from later phonemes (primacy), while the last phoneme, though written at the smallest scale, undergoes no later interference (recency).

Together, these findings suggest that neural systems can solve the binding problem for sequences without dedicated positional slots, by binding each filler to its role through the magnitude and direction of a shared state.

%% file: 6-statements.tex
\section*{Limitations}

Our model conclusions come from a single trained network of one architecture, one hidden size and one language, so we have not shown that a different size or a different recurrent cell learns the same modulation. The size effect reported by \citet{csordas-etal-2024-recurrent} makes it likely that it does not hold everywhere, and testing whether the per-position angle grows toward orthogonality as capacity increases is the natural next step.

The neural claim is bounded more tightly. The evidence is correlational, whereas the model demonstration intervenes on the representation and alters the decoded phoneme, so we show that a magnitude code for within-word position is available in STG, not that it is read out. The rotational component that distinguishes our account from the one we revise is not tested in cortex at all. The identity contrast is coarse, since consonant and vowel separability in STG is already established \citep{mesgarani_phonetic_2014} and our directional effects are small. The dataset is also small next to the model analyses, comprising 1110 delta phonemes drawn from 152 sentences against a vocabulary of 30{,}000 words.

\subsection*{Ethics statement}

The intracranial recordings we analyze were collected for clinical purposes in patients undergoing pre-surgical monitoring for epilepsy, under the protocols and informed consent procedures reported in the original data descriptor \citep{Lakretz_2026}. We performed no new recordings and analyzed only de-identified data.




%% file: 7-appendix.tex
\appendix

\section{Intracranial Recording Details}
\label{app:ieeg}

\paragraph{Dataset.}
We analyze the auditory sentence runs of the intracranial dataset of \citet{Lakretz_2026}, in which 20 participants implanted with depth electrodes for epilepsy monitoring listened to short spoken sentences. The analyzed set comprises 152 sentences, 508 words, and 1618 phonemes. Intracranial electroencephalography resolves cortical activity at millisecond and millimetre scales simultaneously and with a high signal-to-noise ratio, a combination that scalp recordings and functional imaging cannot achieve \citep{parvizi2018}. Its main limitations are the clinical population and the sparse, clinically driven electrode coverage.

\paragraph{Recording and preprocessing.}
For each participant, recordings were notch filtered at 60~Hz, clipped at $5\times$IQR, re-referenced to the median of the channel set and then within each electrode shaft, and screened automatically for bad channels. High-gamma activity, a proxy for local population firing, was computed as the analytic amplitude of the 70--150~Hz band using the Hilbert transform. Epochs were cut from $-0.5$ to $1.5$~s around sentence onset and z-scored against the $[-0.5, 0]$~s baseline.

\paragraph{Electrode selection.} Contacts were assigned to a region by their MNI coordinate under the AAL3 atlas, taking Temporal\_Sup in both hemispheres. Responsiveness was tested with a one-sample cluster permutation test of (response $-$ baseline) $> 0$, with baseline $[-0.3,-0.01]$~s, response $[0,1]$~s, a cluster-forming threshold of $p<0.05$ one-tailed, 1000 sign flips and a cluster threshold of $p<0.01$. Each channel's delay is the onset of its earliest significant cluster.
Of the 108 superior temporal gyrus (STG) electrodes, we retain the 59 that respond significantly to sentence onset (cluster permutation test against baseline). These come from 9 of the 20 participants. Each electrode's signal is shifted by its own response latency (median 43~ms).

\paragraph{Cohort surprisal.} The lexicon is the CMU Pronouncing Dictionary \citep{cmudict} augmented with every pronunciation actually presented, so that no observed prefix is unseen, with add-$0.5$ smoothing applied to the cohort continuation counts.

\section{Intervention Training and Dataset Details}
\label{app:training}

The encoder and decoder each use one LSTM layer with hidden size $d = 128$, so the concatenated encoder state $z = [h; c]$ has $2d = 256$ dimensions.

\paragraph{Settings.} In addition to the two choices of target reported in the main text, counterfactual and word, we test a third setting in which the source sequence the state is taken from and the target sequence are both real English words differing at a single phoneme (Table~\ref{tab:results}).

\paragraph{Optimization.} Interventions are trained with Adam, minimizing the cross-entropy between the target sequence and the decoder output produced from the edited state. Training uses early stopping on validation accuracy, halting after about 25 epochs on average. Unless stated otherwise, appendix comparisons are point estimates from a single trained model, with the dataset fixed across runs by a shared seed, whereas Table~\ref{tab:results_main} in the main text reports means over cross-validation seeds.

\paragraph{Embeddings.} The phoneme embeddings that seed the difference $(x_{\text{new}} - x_{\text{old}})$ in Equation~\ref{eq:intervention} are squashed by a $\tanh$ before differencing. When the embedding is initialized from the repetition model, whose embedding dimension is $d = 128$, it is duplicated by concatenation with itself to match the $2d = 256$-dimensional state, which adds no trainable parameters. Held fixed, the mean-$\Delta z$ and base-model embeddings are indistinguishable, and training either jointly with $\phi$ raises accuracy well above both (Table~\ref{tab:embed}). Which initialization is best once trained is not stable across parameterizations, so we draw nothing from that ordering and use the mean-$\Delta z$ initialization throughout for comparability.

\begin{table}[h]
\centering
\begin{minipage}[t]{0.52\textwidth}
\centering
\caption{Effect of phoneme embedding initialization and joint training, for the spiral method on state $z$ with the counterfactual as target.}
\label{tab:embed}
\small
\renewcommand{\arraystretch}{1.15}
\begin{tabular}{l c c}
\toprule
Embedding init. & Joint train & Accuracy \\
\midrule
\multirow{2}{*}{mean $\Delta z$}
 & \checkmark & 0.720 $\pm$ 0.011 \\
 & \ding{55}  & 0.623 $\pm$ 0.010 \\
\midrule
\multirow{2}{*}{base-model emb.}
 & \checkmark & \textbf{0.807} $\pm$ 0.009 \\
 & \ding{55}  & 0.628 $\pm$ 0.008 \\
\midrule
from scratch
 & \checkmark & 0.750 $\pm$ 0.009 \\
\bottomrule
\end{tabular}
\end{minipage}
\hfill
\begin{minipage}[t]{0.44\textwidth}
\centering
\caption{Intervention accuracy for the main methods in the real-to-real setting, where both source and target sequence are real words from train set. Embeddings are initialized from the mean $\Delta z$ and trained jointly. The unconstrained method is the low-rank form at rank 8.}
\label{tab:results}
\small
\renewcommand{\arraystretch}{1.15}
\begin{tabular}{l c}
\toprule
Method &  Accuracy \\
\midrule
Unconstrained & 0.995 \\
Spiral  & 0.996 \\
Onion & 0.994 \\
DAS   & 0.568\\
DAS bigram & 0.704\\
\bottomrule
\end{tabular}
\end{minipage}
\end{table}

\section{Phonotactic Legality and Intervention Success}
\label{app:phonotactics}

Counterfactual pairs are built by replacing a phoneme at random, so many of the resulting sequences are ones no English word could contain. We define two filters over them. The consonant-vowel filter discards substitutions that violate the consonant-vowel patterns seen in training, and the n-gram filter discards those creating phoneme bigrams or trigrams absent from real-word training sequences. Crossing the two gives the design reported here.

Artificial token sequences are unconstrained, so models evaluated on synthetic data never encounter phonological rules. In natural language, substituting a phoneme at random frequently generates strings that violate the constraints of the language. We therefore asked whether intervention success depends not only on the geometry of the state vector but also on whether the edited state lands in a phonotactically valid region of the decoder's representation space.
We filtered counterfactual targets by structural validity (Table~\ref{tab:filters}). With no filter, intervention accuracy was $0.722$. Enforcing consonant-vowel structure alone raised it to $0.743$, whereas rejecting unattested n-grams had a considerably larger effect, reaching $0.772$ for bigrams and $0.866$ for trigrams. The n-gram filters therefore contributed substantially more than the local CV constraint, which shows that the structure governing what the decoder can read out extends beyond immediately neighbouring phonemes. The two constraints were also largely redundant, since CV added to the bigram filter ($0.772$ to $0.790$) but nothing to the trigram filter ($0.866$ in both cases), as expected if trigram statistics already capture the CV alternation. Because random sequences lack these higher-order constraints, this dependence cannot appear in the synthetic setting where the onion hypothesis was originally derived.

\begin{table}[h]
\centering
\caption{Effect of each phonotactic filter and their combinations, for the spiral model with trained mean-$\Delta z$ embeddings. CV enforces the consonant-vowel patterns of real words and N-gram rejects unattested bigrams (2) or trigrams (3).}
\label{tab:filters}
\begin{tabular}{c c c}
\toprule
CV & N-gram & Accuracy \\
\midrule
\rowcolor{gray!10}
 \ding{55}  & \ding{55} & 0.722 $\pm$ 0.009\\
 \checkmark & \ding{55} & 0.743 $\pm$  0.011 \\
 \rowcolor{gray!10}
 \ding{55}  & 2         & 0.772 $\pm$  0.005 \\
 \checkmark & 2         & 0.790 $\pm$  0.006 \\
 \rowcolor{gray!10}
 \ding{55}  & 3         & \textbf{0.866 $\pm$ 0.005} \\
 \checkmark & 3         & 0.866 $\pm$ 0.010 \\
\bottomrule
\end{tabular}
\end{table}

\section{Interchange Intervention Details}
\label{app:das}

The interchange baseline of Section~\ref{sec:methods} follows \citet{csordas-etal-2024-recurrent}, a variant of distributed alignment search \citep{geiger_das_2024, wu2023boundless}. Each example supplies a base sequence $\mathrm{b}$, whose state is edited, and a source sequence $\mathrm{s}$, whose state is copied from.

\begin{table}[h]
\centering
\caption{Data for replacing the phoneme at position $p = 3$ of \emph{plant} with /IH/. Bold marks the edit, underlining marks the window on which the source must agree with the target, which widens with $k$. Every other source position is resampled. The additive method needs no source sequence.}
\label{tab:examples}
\small
\setlength{\tabcolsep}{6pt}
\renewcommand{\arraystretch}{1.15}
\begin{tabular}{l l l l}
\toprule
Method & Base & Source & Target \\
\midrule
Additive     & \texttt{P L \textbf{AE} N T} & \textemdash                              & \texttt{P L \textbf{IH} N T} \\
DAS, $k = 1$ & \texttt{P L \textbf{AE} N T} & \texttt{K R \underline{\textbf{IH}} S P} & \texttt{P L \underline{\textbf{IH}} N T} \\
DAS, $k = 2$ & \texttt{P L \textbf{AE} N T} & \texttt{F \underline{L \textbf{IH} N} CH} & \texttt{P \underline{L \textbf{IH} N} T} \\
\bottomrule
\end{tabular}
\end{table}

The hypothesis under test is that each position occupies a linear subspace of the encoder state. The subspaces need not be axis-aligned, so the method learns an orthogonal $\mathbf{R} \in \mathbb{R}^{2d \times 2d}$ that rotates $z$ into a basis where they are, together with an assignment matrix $\mathbf{A} \in \{0,1\}^{2d \times (V+1)}$ that gives each rotated dimension to one of $V$ positional variables or to a final column of dimensions that are never intervened on. Subspace sizes are therefore learned rather than fixed. $\mathbf{A}$ is the row-wise hard Gumbel-softmax of a real-valued $\hat{\mathbf{A}}$ with a straight-through estimator, penalized toward the final column.

Let $S$ be the variables the edit touches and $\mathbf{m} = \sum_{v \in S} \mathbf{A}[\cdot, v]$ the dimensions they own. The intervention replaces those dimensions with the source's and rotates back,
\begin{equation}
\begin{aligned}
\hat{\bar{z}} &= \mathbf{m} \odot \mathbf{R} z^{\mathrm{s}} + (\mathbf{1} - \mathbf{m}) \odot \mathbf{R} z^{\mathrm{b}}, \\
\hat{z} &= \mathbf{R}^{\top} \hat{\bar{z}}.
\end{aligned}
\label{eq:das}
\end{equation}
$\mathbf{R}$ and $\hat{\mathbf{A}}$ are trained through the frozen decoder in the same way as $\phi$. A variable spans $k$ consecutive positions, giving $V = P - k + 1$ of them for a maximum sequence length $P$, and $S$ holds every variable whose window covers the edited position. Unlike Equation~\ref{eq:intervention}, which sees only the position and the two phonemes, Equation~\ref{eq:das} sees a full encoder state, so the source has to be built to leak no context, as described in Section~\ref{sec:data}.

\section{Additional Modulation Parameterizations}
\label{app:methods}

\begin{figure}[!t]
    \centering
    \includegraphics[
    page=6,
    width=1\linewidth,
    trim=0cm 12cm 0cm 0cm,
    clip
]{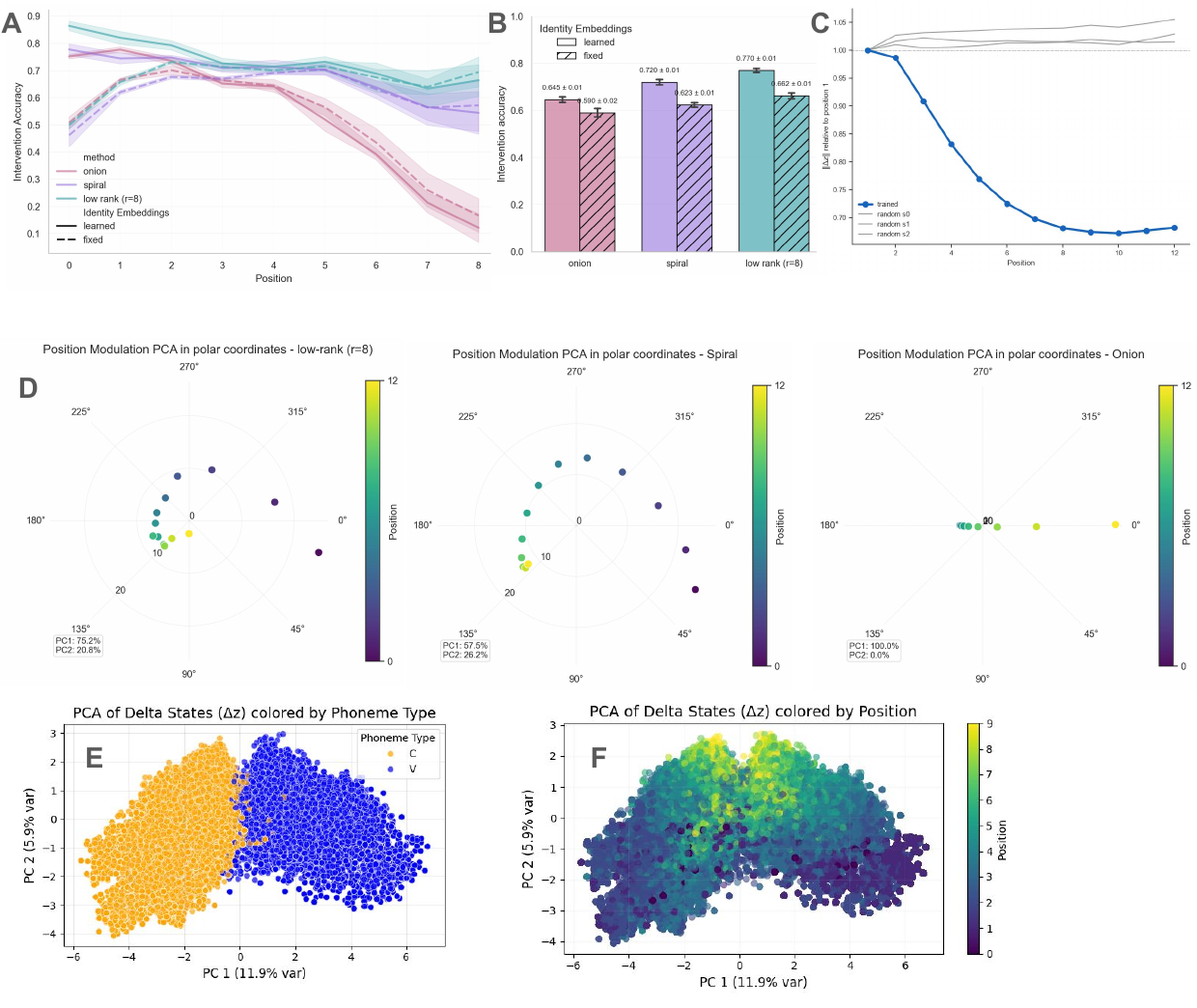}
   \caption{
   \textbf{(A, B)} Intervention accuracy by position for each method, including the unconstrained low-rank-8 method. The spiral method is competitive with the unconstrained method using far fewer parameters.
   \textbf{(C)} Mean $\|\Delta z\|$ by position for the trained repeat model and three randomly initialised encoders of the same architecture, each curve normalised to its own position-1 value (untrained weights sit at an arbitrary scale, so only the shape is comparable). The trained profile decays; the untrained ones are flat. The decay in Figure~\ref{fig:descriptive}A is learned, is not a generic consequence of LSTM dynamics, but a learned modulation that carries position.
   \textbf{(D)} Learned modulation vectors $\phi$ for the main methods in PCA space. The rotation is present in the unconstrained and spiral methods and absent from the onion method.
   \textbf{(E)} Consonants and vowels separate in PCA space.
   \textbf{(F)} Earlier positions are more separated while later positions converge toward the center, which explains why fixed embeddings underperform at early positions.}
    \label{fig:appendix-model}
\end{figure}

Beyond the onion and spiral methods of Section~\ref{sec:methods}, we evaluate the parameterizations below. Throughout, $m = 2d = 256$ is the state dimension and $p$ is the position of the edit.

\subsection{Unconstrained Methods}

\emph{Unconstrained} parameterizations store $\phi(p)$ independently for each position and impose no relationship between positions. They give the most flexible fit and serve two roles, an upper bound on what the additive edit can achieve, and a diagnostic, since inspecting the learned vectors shows whether the position transformation follows a regular pattern.

\paragraph{Per-position.} Each position holds its own vector,
\begin{equation}
\phi(p) = \Phi_{p}, \qquad \Phi \in \mathbb{R}^{P \times 2d},
\end{equation}
where $P$ is the maximum sequence length. This is the most flexible form and uses $P \times 2d$ parameters.

\paragraph{Low-rank.} To keep positions independent while reducing parameters, the modulation matrix is factored,
\begin{equation}
\phi(p) = C_{p} B^\top, \qquad C \in \mathbb{R}^{P \times r}, \quad B \in \mathbb{R}^{2d \times r},
\end{equation}
with $r \ll 2d$. The rank $r$ trades flexibility against parameter count, and sharing $B$ across positions also regularizes the fit.

\subsection{Scale-Only Methods}

\paragraph{Linear.} The onion form without its exponential term,
\begin{equation}
\phi(p) = \beta\, p + b, \qquad \beta, b \in \mathbb{R}^{m}.
\end{equation}

\subsection{Scale-and-Rotation Methods}

\paragraph{Lie-group spiral.} A single matrix exponential rotates all coordinates together \citep{Hall2003LieGL},
\begin{equation}
R(p) = \exp(p\, A), \qquad A = W - W^{\top}, \quad A \in \mathbb{R}^{m \times m},
\end{equation}
with $\phi(p) = R(p)\, s(p)$ and $s$ the Decay scale. Skew-symmetry of $A$ makes $R(p)$ orthogonal at every position. This form can couple any pair of coordinates rather than the fixed planes of the spiral method, so it is strictly more expressive, at the cost of $m(m-1)/2$ parameters and a matrix exponential per position.

\subsection{Isolating the Contribution of the Rotation}
\label{app:rotation}

Crossing the scale parameterization with the presence of a rotation separates the contribution of the rotation from that of the scale (Table~\ref{tab:rotation}). Adding the rotary construction to the exponentially decaying scale raises accuracy from $0.64$ to $0.72$, whereas adding the same rotation to the onion scale lowers it from $0.65$ to $0.62$. The rotation therefore helps only when it is applied to a scale that decays monotonically, which is what the spiral hypothesis describes. The onion scale carries a linear term in addition to its exponential one, so its direction already drifts with position, and imposing a further rotation on it appears to overparameterize the transformation rather than add to it.

\begin{table}[h]
\centering
\caption{Intervention accuracy for each combination of scale parameterization and rotation, with the counterfactual as target. The rotation improves the monotonically decaying scale and degrades the onion scale.}
\label{tab:rotation}
\small
\renewcommand{\arraystretch}{1.15}
\begin{tabular}{l c c}
\toprule
& \multicolumn{2}{c}{Rotation} \\
\cmidrule(lr){2-3}
Scale & None & Rotary \\
\midrule
Onion              & 0.65 & 0.62 \\
Exponential decay  & 0.64 & \textbf{0.72} \\
\bottomrule
\end{tabular}
\end{table}

\subsection{Comparison of Parameterizations}

Table~\ref{tab:comparison_methods} compares all parameterizations on the source-to-counterfactual set. Low-rank at rank 4 and 8 is best, slightly above the unconstrained per-position look-up, which is consistent with regularization, since the full look-up has one free vector per position and can overfit the training counterfactuals while a low rank shares parameters across positions and generalizes better. Among the functional methods the spiral form is strongest, above the Lie-group spiral and above the scale-only onion and decay forms, and decay does not improve on onion. On the fully filtered set the simplest baselines, linear and rank 1, remain close to the rest, which reflects how much the filters ease the task.

\begin{table}[h]
\centering
\caption{All parameterizations on the source-to-counterfactual set, with trained and fixed mean-$\Delta z$ embeddings, under the repeat-only filter and under all filters.}
\label{tab:comparison_methods}
\renewcommand{\arraystretch}{1.15}
\begin{tabular}{l c c c c}
\toprule
 & \multicolumn{2}{c}{Trained emb.} & \multicolumn{2}{c}{Fixed emb.} \\
\cmidrule(lr){2-3} \cmidrule(lr){4-5}
Method & Repeat & All filt. & Repeat & All filt. \\
\midrule
\rowcolor{gray!10}
Low-rank-8 & \textbf{0.780} & \textbf{0.898} & 0.678 & 0.807 \\
Per-position & 0.773 & 0.878 & 0.679 & 0.800 \\
\rowcolor{gray!10}
Low-rank-4 & 0.736 & 0.889 & 0.663 & 0.802 \\
Spiral & 0.729 & 0.866 & 0.632 & 0.770 \\
\rowcolor{gray!10}
Low-rank-2 & 0.723 & 0.865 & 0.660 & 0.788 \\
Lie spiral & 0.721 & 0.854 & 0.601 & 0.747 \\
\rowcolor{gray!10}
Linear & 0.658 & 0.794 & 0.590 & 0.739 \\
Decay & 0.651 & 0.807 & 0.599 & 0.707 \\
\rowcolor{gray!10}
Onion & 0.650 & 0.815 & 0.601 & 0.769 \\
Low-rank-1 & 0.635 & 0.768 & 0.554 & 0.715 \\
\bottomrule
\end{tabular}
\end{table}

\section{Which State Carries the Code}
\label{app:state}

The concatenated state $z = [h; c]$ and the cell state $c$ alone give nearly identical accuracy, while editing the hidden state $h$ alone fails completely (Table~\ref{tab:states}). The positional code the intervention exploits therefore lives in the cell state, which is consistent with the cell state carrying the memory the decoder reads.

\begin{table}[h]
\centering
\caption{Intervention accuracy by the state edited, on the source-to-counterfactual set with all filters, the rank-8 method, across embedding settings. When the intervention edits $c$, the decoder receives the source sequence's $h$ together with the intervened $c'$, and the reverse holds for editing $h$.}
\label{tab:states}
\renewcommand{\arraystretch}{1.15}
\begin{tabular}{l c c c c}
\toprule
 & \multicolumn{2}{c}{mean $\Delta z$} & \multicolumn{2}{c}{repetition emb.} \\
\cmidrule(lr){2-3} \cmidrule(lr){4-5}
State & Trained & Fixed & Fixed & Trained \\
\midrule
$z = [h;c]$  & 0.899 & 0.807 & 0.739 & 0.893 \\
$c$ (cell)   & 0.888 & 0.768 & 0.737 & 0.896 \\
$h$ (hidden) & 0.000 & 0.000 & 0.000 & 0.000 \\
\bottomrule
\end{tabular}
\end{table}